\pdfoutput=1

\documentclass[11pt]{article}

\usepackage[table,xcdraw]{xcolor}
\usepackage[]{emnlp2021}
\usepackage{times}
\usepackage{latexsym}
\usepackage{amsmath}
\usepackage{bbm}
\usepackage{graphicx}
\usepackage{multirow}
\usepackage{aliascnt}

\usepackage[T1]{fontenc}

\usepackage[utf8]{inputenc}

\usepackage{microtype}

\title{Temporal Relation Extraction with a Graph-Based Deep Biaffine Attention Model}

\author{Bo-Ying Su \\ University of California, San Diego \\ b1su@ucsd.edu \And
        Shang-Ling Hsu \\ Hong Kong University of Science and Technology \\ shsuaa@connect.ust.hk \AND
        Kuan-Yin Lai \\ National Taiwan University \\ eddy50811@gmail.com \And
        Amarnath Gupta \\ University of California, San Diego \\ a1gupta@ucsd.edu }

\begin{document}
\maketitle
\begin{abstract}
Temporal information extraction plays a critical role in natural language understanding. 
Previous systems have incorporated advanced neural language models and have successfully enhanced the accuracy of temporal information extraction tasks. 
However, these systems have two major shortcomings. 
First, they fail to make use of the two-sided nature of temporal relations in prediction. 
Second, they involve non-parallelizable pipelines in inference process that bring little performance gain.
To this end, we propose a novel temporal information extraction model based on deep biaffine attention to extract temporal relationships between events in unstructured text efficiently and accurately. 
Our model is performant because we perform relation extraction tasks directly instead of considering event annotation as a prerequisite of relation extraction. 
Moreover, our architecture uses Multilayer Perceptrons (MLP) with biaffine attention to predict arcs and relation labels separately, improving relation detecting accuracy by exploiting the two-sided nature of temporal relationships.
We experimentally demonstrate that our model achieves state-of-the-art performance in temporal relation extraction. 
\end{abstract}

\section{Introduction}

Event temporal relation extraction from text is a crucial task in natural language understanding, given its fundamental role in common sense reasoning, information retrieval, and clinical applications, such as disease progression.
The extraction of temporal relations with events comprises several sub-tasks, including the identification of event expressions and temporal relations between events and times, as formally defined in SemEval-2016 \citep{bethard2016semeval}. 
\begin{figure}[t!]
    \centering
    \includegraphics[width=0.49\textwidth]{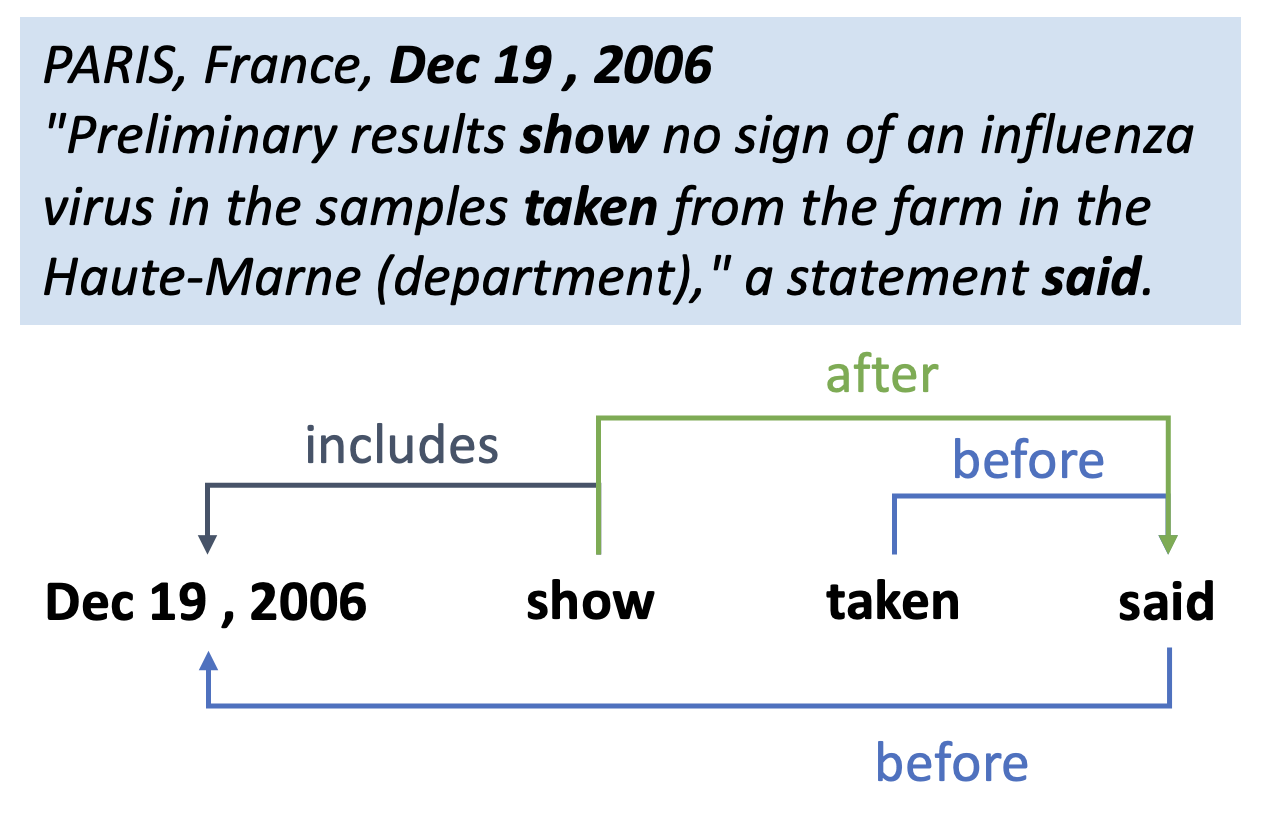}
    \caption{An illustration of temporal relation extraction with events. For simplicity, only the relations between time expressions and events are shown, but in practice, the relations involving other tokens are set to $\textit{NONE}$. Our focus is to extract Event-Event relations in text.}
    \label{fig:example}
\end{figure}
An example of temporal relation extraction with events is shown in Figure \ref{fig:example}, where \textbf{Dec 19, 2006} is the document creation time, and three event instances occur in the sentence: \textbf{show}, \textbf{taken}, and \textbf{said}.
According to the annotations, the time span of the event \textbf{show} $\textit{INCLUDES}$ \textbf{Dec 19, 2006} and is $\textit{AFTER}$ \textbf{said}.
Besides, \textbf{said} happens $\textit{BEFORE}$ \textbf{Dec 19, 2006}, while \textbf{taken} happens even $\textit{BEFORE}$ \textbf{said}.
The other relations between time expressions and event instances may be inferred accordingly or marked as $\textit{VAGUE}$, such as the one between \textbf{show} and \textbf{taken}.


Early studies in event temporal relation extraction rely on human-engineered features \cite{wu2014negation, daume2007frustratingly, blitzer2006domain}, which may experience performance drop upon out-of-domain applications \cite{dligach2017neural}.
Later, a variety of neural models, such as convolutional neural network (CNNs) \cite{dligach2017neural} and long short-term memory neural networks (LSTMs) \cite{tourille2017neural}, outperform handcrafted feature-based models.
Structured learning approaches are also proposed to enhance performance by incorporating more knowledge \cite{ning2017structured, han2019joint}.
Recent research efforts further explore the power of pre-trained Transformer-based models \cite{vaswani2017attention} on temporal relation extraction by proposing extensions and variants \cite{wang2019extracting,ning2019improved,yang2019exploring,han2019joint,han2020domain}.
Latest state-of-the-art performance has been achieved by end-to-end, multi-task approaches that conduct both event annotation and temporal relations extraction jointly \cite{han2019joint, han2020domain, lin2020joint}.
Nonetheless, even if these systems are able to implement joint information extraction, they suffer from performance issues in the inference speed and accuracy, which may be the result of the computationally expensive decoding process and insufficient fusion of information extracted for classification.

To this end, we propose a Graph-Based Deep Biaffine Attention Model. 
Leveraging deep biaffine attention mechanisms and a lightweight decoder module while predicting the relations as a unified graph instead of isolated components, our end-to-end, multi-task model speeds up the inference time for around 39\% and is able to achieve state-of-the-art performance on benchmark datasets of temporal relation extraction with events.

We summarize our contributions below:
\begin{itemize}
    \item We propose the novel application of deep biaffine attention mechanisms in information extraction and demonstrate the benefits of separating arc and relation prediction.
    \item We speedup the decoding process of relation extraction by around 39\% with highly parallelizable pipeline without a significance drop in terms of F1 score.
    \item Our model outperforms the state-of-the-art baselines in temporal relation extraction in benchmark datasets.
\end{itemize}

\section{Problem Specification}
We focus on the problem of extracting relations between events from raw text.
The goal is to generate a Temporal Relation Graph from unstructured text.
A Temporal Relation Graph represents the event-event temporal relationships in a piece of text, with events being the vertices and relations being the edges. Depending on the dataset, the edges in the Temporal Relation Graph may comprise of different set of labels.
Our system is end-to-end, which means that event and relation annotations are only given in the training stage. 
On contrast, during prediction, only raw sentences are given, and the relations are predicted without prior event annotations.
Solving such a problem is useful in real-world settings, as neither annotations of events nor relation data is easily accessible.

\section{Graph-Based Deep Biaffine Attention Model}
\begin{figure}[t!]
    \centering
    \includegraphics[width=0.49\textwidth]{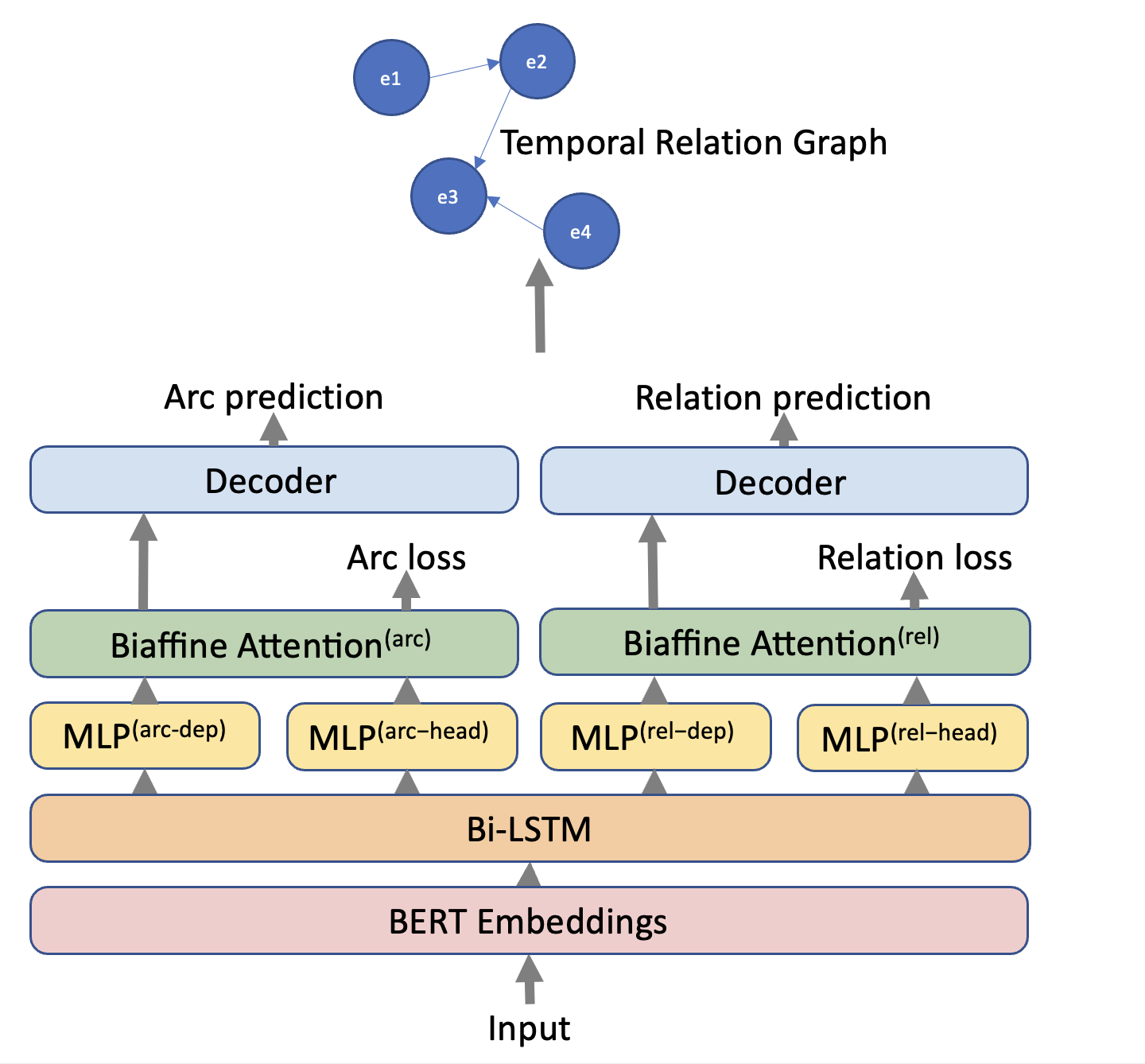}
    \caption{Neural architecture of Graph-Based Deep Biaffine Attention Model.}
    \label{fig:overview_of_proposed_approach}
\end{figure}

In this section, we describe the details of our proposed Graph-Based Deep Biaffine Attention Model.

The input sentences are first tokenized and embeded with a fine-tuned pre-trained BERT model. 
Then, the sequence of tokens is encoded with a Bidirectional-LSTM layer to create a neural representation of the input. 
Two separated Biaffine Attention Modules, ARC and REL, will then predict arcs and relation labels between each pair of tokens based on the neural representation respectively. These two pieces of information correspond to the edges and labels in the Temporal Graph.
Finally, the decoder combines the most probable arc and temporal relation labels to make predictions of the temporal relations between each pair of tokens.
We perform multi-task learning on our model. In other words, the losses of the ARC and REL module are combined to a total loss and are optimized jointly using cross-entropy loss. 

Although we do utilize gold event span information during training, event information is not used during inference, which eliminates the requirement of event annotations and furthers the potential of the method to be put into production.

\subsection{Preprocessing}
In this section, we describe how we preprocess the raw annotations from the dataset.

Since the TimeBank-Dense dataset (TB-Dense) only annotates event relations across sentences that are not further than two sentences away \cite{cassidy2014annotation}, we apply a sliding window to capture the tokens in every two consecutive sentences and the event-event relations among them.

Then, we take a data augmentation step on the dataset.
Consider $e_{1} \subseteq X$ and $e_{2} \subseteq X$ to be two annotated events, each containing one or more tokens from the dataset $X$, a relation tuple $(\mathrm{x}_{1}, \mathrm{x}_{2}, r)$ containing two tokens $\mathrm{x}_1, \mathrm{x}_2 \in X$ and a relation label $r$ is generated if and only if $\mathrm{x}_{1} \in  e_{1}$ and $\mathrm{x}_{2} \in  e_{2}$ and there is TLINK annotation between $e_{1}$ and $e_{2}$ in the dataset. This step effectively produces a dense graph among two event spans if there is a temporal relation between them. \par
All the inverse relations that could be inferred immediately from the dataset is also added to the training data.

\subsection{BERT Fine-tuning}
Similar to previous studies that utilizes pre-trained BERT model, we leverage the Masked Language Modeling target with implementation from huggingface \cite{wolf2019huggingface} to fine-tune the BERT-base model on each training corpus. 

\subsection{ARC and REL module}

Our Graph-Based Deep Biaffine Attention Model makes two predictions for any given text: arc (ARC) and relation (REL). In this section, we describe how ARC and REL modules work. \par

\medskip
\noindent\textbf{ARC}\quad The purpose of the ARC module is to predict the existence of an arc between any pair of tokens. Additionally, the ARC module output also give us information about the direction of the arc, although it is not used. 
The motivation of the ARC module is to convert the problem of detection the existence of a temporal relationship between tokens to a prediction of a directed edge between a pair of tokens. This effectively gives the model more information about the relations labels than simply telling it if there is an arc or not.

For every text input, we create a binary two-dimensional matrix $\textit{ARC}_{i,j} \in \{0,1\}^{n \times n}$, where $n$ denotes the sequence length, which is exactly the adjacency-list representation of the graph.
$\textit{ARC}_{i,j} = true$ if only if there is an arc that points from $\mathrm{x}_{i}$ to $\mathrm{x}_{j}$, where $\mathrm{x}_{i}$ denotes the $i^{th}$ token in the sentence.

For each pair of two-sided temporal relations, such as $\{\textit{AFTER},\textit{BEFORE}\}$ and $\{\textit{INCLUDE},\textit{IS\_INCLUDED}\}$, we arbitrary choose one side to be the source and the other to be the sink.
Consequently, if there is a two-sided temporal relationship between a pair of token $\mathrm{s}_{i}$ and $\mathrm{s}_{j}$, we have either $\textit{ARC}_{i,j} = true$ or $\textit{ARC}{j,i} = true$, but not both.

For those temporal relations that does not have their counterparts, such as $\textit{SIMULTANEOUS}$ and $\textit{VAGUE}$, we simply create two edges between them in $\textit{ARC}$. For example, if there is a token-relation pair $( t_{i} ,t_{j} , \textit{SIMULTANEOUS})$, the model will set $\textit{ARC}_{i,j} = \textit{ARC}_{j,i} = true$

A Biaffine attention operator is used to predict the $\textit{ARC}$ matrix, which we will discuss in the Biaffine Attention section below. \par

The cross-entropy loss of the ARC module is calculated as follows:
\begin{equation}
\begin{split}
\mathcal{L}_{arc}&  =loss\left( ARC^{gold} ,\mathbf{s}^{( arc)}\right) \\
& =-\frac{1}{N^{2}}\sum _{i,j} ARC_{i,j}^{gold} \cdot log\left(\mathbf{s}_{i,j}^{( arc)}\right)+\\
 & \left( 1-ARC_{i,j}^{gold}\right) \cdot log\left( 1-\mathbf{s}_{i,j}^{( arc)}\right)
\end{split}
\end{equation}

Since our model does not have any event information during inference, the ARC module is trained to identify possible arcs among all token pairs. This is accomplished by sampling 50\% of token pairs randomly in the input sentence during training.

\medskip
\noindent\textbf{REL}\quad The purpose of the REL module is to predict the temporal relation label between any pair of token. Here is a list of temporal relations that is annotated in the two datasets that we used for evaluation.

\begin{table}
\centering
\begin{tabular}{cc}
\hline
\textbf{TB-Dense} & \textbf{MATRES}\\
\hline
BEFORE & BEFORE \\
AFTER & AFTER \\
SIMULTANEOUS & SIMULTANEOUS \\ 
VAGUE & VAGUE \\ 
INCLUDE & - \\
IS\_INCLUDED & -  \\ 
\hline
\end{tabular}
\caption{Temporal Relation Set annotated in evaluation datasets.}
\label{tab:accents}
\end{table}

For every text input, we create a positive integer two-dimensional matrix $\textit{REL} \in \{1,2,...,n\}^{m \times m}$, where $\textit{REL}_{i,j}$ stores the temporal relation labels if there is an arc that points from $\mathrm{x}_{i}$ to $\mathrm{x}_{j}$, $m$ denotes the sequence length, and $n$ denotes the number of relations.
Due to the two sided nature of temporal relations, we always have $\textit{REL}_{i,j} = {INV}(\textit{REL}_{j,i})$. The inverse mapping is provided in Table \ref{tab:inverse}.

The cross-entropy loss of the REL module is calculated as follows:
\begin{equation}
    \begin{split}
        \mathcal{L}_{rel} & =loss\left( REL^{gold} ,\mathbf{s}^{( rel)}\right) \\ 
        & =\sum _{i,j<i} -log\left(\frac{exp\left(\mathbf{s}^{( arc)}_{i,j,REL_{i,j}^{gold}}\right))}{\sum _{k} exp\left(\mathbf{s}_{i,j,k}^{( rel)}\right)}\right)
    \end{split}
\end{equation}

\begin{table}
\centering
\begin{tabular}{cc}
\hline
\textbf{$REL$} & \textbf{$INV(REL)$}\\
\hline
BEFORE & AFTER \\
AFTER & BEFORE \\
SIMULTANEOUS & SIMULTANEOUS \\ 
VAGUE & VAGUE \\ 
INCLUDE & IS\_INCLUDED \\
IS\_INCLUDED & INCLUDE  \\ 
\hline
\end{tabular}
\caption{Inverse mapping between two Temporal Relation labels.}
\label{tab:inverse}
\end{table}

Note that since $\textit{REL}_{i,j}$ provides the exact same information as $\textit{REL}_{j,i}$ (the later is just the inversion of the former), the loss function of REL only takes account of the upper-half of the $\textit{REL}_{j,i}$ tensor.

\medskip
\noindent\textbf{Conclusion}\quad By predicting both ARC matrices and REL tensors, we are able to construct a token-level temporal relation graph based on the input text. The ARC module predicts the existence of an edge in the token-level temporal graph, whereas the REL module gives us the fine-grained temporal relation labels. 

\subsection{Sentence Encoder}
Similar to prior temporal relation extraction systems \cite{han2019joint,han2020domain,ning2019improved}, we use a BiLSTM layer to encode the token embeddings from the BERT model. The encoded sentences are then passed into four independent Multilayer Perceptrons (MLP), which we will describe below.

\subsection{MLP and Biaffine Attention Layer}
The purpose of this Deep Biaffine Attention layer is to predict the ARC matrices and REL tensors which are mentioned earlier.
Deep Biaffine Attention Architecture is first used to predict dependency edges \cite{dozat2016deep}. The predicted edges and labels are used to construct the dependency tree for a sentence, producing superior parsing accuracy and performance. 
In this work, we modified the Deep Biaffine Attention mechanism to predict directed edges and associated labels for constructing Temporal Relation Graphs. \par

\medskip
\noindent\textbf{Predicting ARC}\quad The goal here is to make a prediction of the ARC matrix. 
For each sentence encoding output from the BiLSTM layer, we first pass it through two Multilayer Perceptrons (MLP) to reduce its hidden dimension in half, creating two vectors, $\mathbf{h}_{i}^{( arc-dep)}$ and $\mathbf{h}_{j}^{( arc-head)}$.
The reason of creating two separated vectors is to allow the two MLPs to be trained to distill head and dependent information for the Biaffine Attention Layers from the BiLSTM output, reducing its vector dimensions.
Here, we use the term "head" to represent an encoded token being the source of other tokens (pointing outwards),
whereas "dep" represents a token being a child of another token.

With $\mathbf{h}_{i}^{( arc-dep)}$ and $\mathbf{h}_{j}^{( arc-head)}$, we apply the Biaffine operator to compute the attention of all tokens to every other tokens, creating a two dimensional matrix $\mathbf{s}_{j,k}^{( arc)} \in \mathbbm{R}^{m \times m}$, where $m$ denotes the sequence length: \par

\begin{equation}
\mathrm{\mathbf{s}_{i,j}^{( arc)} =\mathrm{Biaffine}\left( \mathbf{h}_{i}^{( arc-dep)} ,\mathbf{h}_{j}^{( arc-head)}\right)}
\end{equation}
\begin{equation}
\mathrm{Biaffine}(\mathbf{y_{1}} ,\mathbf{y_{2}}) =\underbrace{\mathbf{y_{1}^{\top } Uy_{2}}}_{Bilinear} +\underbrace{\mathbf{W}(\mathbf{y_{1} \circ y_{2}}) +\mathbf{b}}_{Linear}
\end{equation}\label{eq:biaffine-arc}
\par

Here, $\mathbf{U}$ ,$\mathbf{W}$ and $\mathbf{b}$ are a $m\times m$ tensor, $1 \times 2m$ matrix and a bias vector, where
$m$ is the sequence length of the given text (same as the length of the BiLSTM output).

The intuition here is to model both the prior probability of a token $i$ having any dependents ($\mathbf{W}(\mathbf{y_{i}}$) and the likelihood of $i$ receiving a specific dependent $j$ ($\mathbf{y_{i}^{\top } Uy_{j}}$). In both cases, the higher probability means that we are more confident that there is an arc connected to $i$
$\mathbf{s}\mathnormal{_{i,j}^{( arc)}}$ is then served as the prediction of $\textit{ARC}(i,j)$.

\medskip
\noindent\textbf{Predicting REL}\quad The goal of this module is to assign every possible arcs between tokens a temporal relation label. Here, we leverage a Fixed-class Biaffine classifier.

Similar to predicting the ARC matrix, we feed the sentence encoding output from the BiLSTM layer through two Multilayer Perceptrons (MLP) to reduce its hidden dimension in half, creating two vectors $\mathbf{h}_{i}^{( arc-dep)}$ and $\mathbf{h}_{j}^{( arc-head)}$.
Then, the Biaffine Attention operator is used to compute the likelihood $\mathbf{s}_{i,j}^{(rel)} \in \mathbbm{R}^{m \times m \times n}$ of a token pair $x_i$, $x_j$ connected by a temporal relation label $k$, where $m$ denotes the sequence length, and $n$ denotes the number of relation labels.

\begin{equation}
\mathrm{\mathbf{s}_{i,j}^{( rel)} =Biaffine\left( \mathbf{h}_{i}^{( rel-dep)} ,\mathbf{h}_{j}^{( rel-head)}\right)}
\end{equation}
\begin{equation}
\mathrm{Biaffine}(\mathbf{y_{1}} ,\mathbf{y_{2}}) =\underbrace{\mathbf{y_{1}^{\top } Uy_{2}}}_{Bilinear} +\underbrace{\mathbf{W}(\mathbf{y_{1} \circ y_{2}}) +\mathbf{b}}_{Linear}
\end{equation}

Here, $\mathbf{U}$ ,$\mathbf{W}$ and $\mathbf{b}$ are a $m\times l\times m$ tensor, $l\times 2m$ matrix and a bias vector where
$m$ is the sequence length of the given text (same as the length of the Bi-LSTM output) and $l$ is the number of temporal relation labels.

This construction allows the optimizer to learn the prior probability of a certain label being assigned to any pair of tokens, the likelihood of a head token taking a particular relation and how probable a relation is assigned given a head and a specific dependent.
$\mathbf{s}\mathnormal{_{i,j}^{( rel)}}$ is then served as the prediction of $\textit{REL}(i,j)$.


The Biaffine Attention mechanism allows us to model the temporal relation extraction task directly, instead of introducing additional non-linear activation functions to our model.

\subsection{Loss function}
In our Multi-task model that trains ARC and REL jointly, we compute the loss by simply adding ARC loss with REL loss.
$\mathcal{L}_{arc-rel} =\mathcal{L}_{arc} +\mathcal{L}_{rel}$

\subsection{Decoder}
Our Event Module uses the same implementation as (ref??), with two Linear layers and a tanh activation function in between
The output of the Event module is an matrix of logits of length 2.

The ARC decoder predicts the arc between each pair of tokens $x_i$, $x_j$ as 
\begin{equation}
ARC_{pred}(i,j) = \mathbbm{1}_{\mathbbm{R}+} (\mathbf{s}_{i,j}^{( arc)}),
\end{equation}
where $\mathbbm{1}$ is an indicator function.

The Relation Decoder Module predicts the relation label between each pair of tokens. 
The first part is predicting the existence and the direction of an arc between two tokens $x_i$, $x_j$.
As
${LABEL}_{pred}( i,j)$ and ${REL}_{pred}( i,j)$ are defined as follows:

\begin{equation}
    {LABEL}_{pred}( i,j) =\underset{k}{argmax}( \mathbf{s}_{i,j,k}^{(rel)})
\end{equation}
\begin{equation}
    a = {ARC}_{pred}( i,j)+{ARC}_{pred}( j,i)
\end{equation}
\begin{equation}
    {REL}_{pred}( i,j) =
    \begin{cases} 
      0 & \mathrm{if}\quad a=0 \\
      {LABEL}_{pred}( i,j) & \mathrm{o.w.}
   \end{cases}
\end{equation}

\section{Experiment Setup}
This section describes the details about the two datasets and evaluation metrics that we used to train and evaluate our system. 
We use our workstation 

\subsection{Dataset}
We use TimeML-annotated \cite{pustejovsky2004specification} TimeBank-Dense dataset (TB-Dense) and Multi-Axis Temporal Relations for Start-points (MATRES) dataset to evaluate the performance of our system.
Both TB-Dense and MATRES dataset provide event annotations and relations between events.
High-quality temporal relation data has always been scarce resource, due to its challenging annotation process. Annotators often overlook some relations that can be implied from text, but hard to recognize at first glance, resulting in a low inter annotator agreement.

\begin{table}[]
\centering
\begin{tabular}{|c|c|c|}
\hline
\textbf{}        & TB-Dense        & MATRES      \\ \hline
\multicolumn{3}{|c|}{\textbf{\# Documents}}      \\ \hline
Train            & 22              & 183         \\ \hline
Dev              & 5               & -           \\ \hline
Test             & 9               & 20          \\ \hline
\multicolumn{3}{|c|}{\textbf{\# Relation Pairs}} \\ \hline
Train            & 4032            & 6332        \\ \hline
Dev              & 629             & -           \\ \hline
Test             & 1427            & 827         \\ \hline
\end{tabular}
\caption{Dataset splits of TB-Dense and MATRES.}
\label{tab:splits}
\end{table}

\begin{table}[]
\centering
\begin{tabular}{|c|c|c|}
\hline
\multicolumn{1}{|l|}{} & \textbf{TB-Dense} & \textbf{MATRES} \\ \hline
\underline{B}EFORE                 & 384                     & 417             \\
\underline{A}FTER                  & 274                     & 266             \\
\underline{I}NCLUDES               & 56                      & -               \\
\underline{I}S\_\underline{I}NCLUDED           & 53                      & -               \\
\underline{S}IMULTANEOUS           & 22                      & 31              \\
\underline{V}AGUE                  & 638                     & 113             \\ \hline
\end{tabular}
\caption{Relations in TB-Dense and MATRES. Acronyms of the relations are underlined.}
\label{tab:stats-rel}
\end{table}

\subsubsection{TB-Dense}
The TB-Dense dataset mitigates such an issue by forcing the annotators to consider each pair of events that co-exist in sentences that is no father than two sentences away. The result is a dataset that contains approximately 10x more relations per document compared to the original Timebank dataset \cite{cassidy2014annotation}. 

\subsubsection{MATRES}
Recent work by \citet{NingWuRo18} further improves the quality of temporal annotations by enforcing a multi-axis annotation scheme, improving the inner-annotator agreements significantly.

We use TB-Dense and MATRES datasets to evaluate our model. The data statistics are provided in Table \ref{tab:stats-rel}.

\subsection{Evaluation Metrics}
To be consistent with previous research and the baseline models, we adopt the same standard micro-average scores for TB-Dense dataset. NONE relations are excluded from the calculations, following the convention of Information Extraction tasks.

For MATRES dataset, we also use the same micro-average metric as the baseline models.

Since the predictions from the proposed model is at token-level, and it does not predict event spans, we use the first token in every event as representing the entire event. In other words, all relations that is associated to a event in the gold annotation must exist as relations from/to the first token of the event to be considered correct.

The code for evaluation is provided in fmp/utils/metric.py
\section{Baseline Models}
We pick the state-of-the-art methods from Feature-Based Systems and Neural Systems to be our baselines.
\subsection{Feature-Based Systems Baseline}


For the TB-Dense dataset, we use CAEVO \cite{mirza-tonelli-2016-catena} as our feature-based system baseline. CAEVO is a rule-based system with MaxEnt classifier, utilizing linguistic features. It is the best performing feature-based temporal relation extraction system for TB-nse.
For the MATRES dataset, we use the end-to-end system Cog-CompTime as our feature-based system baseline.

\subsection{Neural Systems Baseline}
We use the system by described by \citet{han2020domain} as the baseline of the TB-Dense dataset, as it reports the best end-to-end performance to date. For MATRES, we use the end-to-end system by \citet{han2019joint} as our baseline model.

\section{Results}

\begin{table*}[t]\centering
\begin{tabular}{|c|c|c|c|c|c|c|}
\hline
\multicolumn{1}{|l|}{\multirow{3}{*}{}} & \multicolumn{3}{c|}{\textbf{TB-Dense}}                                      & \multicolumn{3}{c|}{\textbf{MATRES}}                                              \\ \cline{2-7} 
\multicolumn{1}{|l|}{}                  & \textbf{P}                & \textbf{R}                & \textbf{F1}               & \textbf{P}                & \textbf{R}                & \textbf{F1}               \\ \hline
CAVEO \citep{mirza-tonelli-2016-catena}& 43.8                      & 35.7                      & 39.4                      & 62.7                      & 58.7                      & 60.6                      \\
\citealp{han2019joint} & \multicolumn{1}{c|}{52.6} & \multicolumn{1}{c|}{46.5} & \multicolumn{1}{c|}{49.4} & \multicolumn{1}{c|}{59.0} & \multicolumn{1}{c|}{60.2} & \multicolumn{1}{c|}{59.6} \\
\citealp{han2020domain} & \textbf{53.4}                      & 47.9                      & 50.5                      & 42.6                         & 52.6                         & 46.5                         \\
\hline
\textbf{Graph-based Biaffine}           & 50.6                     & \textbf{51.3}                      & \textbf{50.9}                      & \textbf{65.8}                      & \textbf{66.8}                      & \textbf{66.3}                      \\ \hline
\end{tabular}
\caption{Comparison of our Graph-based Biaffine Model with the baselines.}
\label{tab:mainresult}
\end{table*}

Table \ref{tab:mainresult} contains our main results. 
In this section, we discuss the performance on the two datasets respectively.

\subsection{TB-Dense}
The proposed Graph Biaffine model outperforms the feature baseline by 11.5\% and surpasses the neural baseline by 0.4 in the TB-Dense dataset. When looking at the score breakdown per label in Table \ref{tab:perform-end-to-end}, the proposed Graph-based Deep Biaffine model performs much better in predicting \textit{INCLUDE} and \textit{IS\_INCLUDED} temporal relation labels, while falling short on predicting \textit{BEFORE} and \textit{AFTER}. We show that with a dedicated ARC module that detects arcs in the Temporal Graph instead of treating not having a relation merely as \textit{NONE} label, the relation classifier could be much better at dealing wih imbalanced data. In the TB-Dense dataset, \textit{BEFORE} and \textit{AFTER} relations account for 46.1\% of all temporal relation annotations, while \textit{INCLUDE} and \textit{IS\_INCLUDED} account for only 7.6\% percent in total.
Additionally, the F1 score of \textit{VAGUE} relation improved ~7.9\%. We speculate this is due to our model being superior in distinguishing whether a relation exist between two tokens or not. 
\textit{SIMULTANEOUS} remains to be the only label that is not predicted at all, probably due the the lack of samples in the training set (only 22 according to Table \ref{tab:stats-rel}).

\subsection{MATRES}
Our proposed model also outperforms baseline models in MATRES dataset by a margin of 5.7\%. Since every event is a single verb in MATRES, the proposed model does not have to ensure all relations that is associated to a event to be present in every single token within its span. This explains why the improvement margin larger compared to that of TB-Dense dataset.

\begin{table}[]
\centering
\begin{tabular}{|c|c|c|c|c|c|c|}
\hline
\multicolumn{1}{|l|}{\multirow{2}{*}{}} & \multicolumn{3}{c|}{\textbf{Baseline}} & \multicolumn{3}{c|}{\textbf{Ours}} \\ \cline{2-7} 
\multicolumn{1}{|l|}{}                  & \textbf{P}     & \textbf{R}     & \textbf{F1}     & \textbf{P}    & \textbf{R}    & \textbf{F1}   \\ \hline
\textbf{B}                              & 58.6           & 55.7           & 57.1            & 54.3          & 37.4          & 44.2          \\
\textbf{A}                              & 67.8           & 51.5           & 58.5            & 57.3          & 35.2          & 43.5          \\
\textbf{I}                              & 8.3            & 1.8            & 2.9             & 12.1          & 5.4           & 7.48          \\
\textbf{II}                             & -              & -              & -               & 15.2          & 7.6           & 10.1          \\
\textbf{S}                              & -              & -              & -               & -             & -             & -             \\
\textbf{V}                              & 47.6           & 51.4           & 49.4            & 50.7          & 66.0          & 57.3          \\ \hline
\textbf{Avg}                            & \textbf{53.4}           & 47.9           & 50.5            & 50.6          & \textbf{51.3}          & \textbf{50.9}          \\ \hline
\end{tabular}
\caption{Comparison of the performance of different labels in the end-to-end settings on TB-Dense dataset. The baseline system The score breakdown is proposed by \citet{han2020domain}. Acronyms of the labels correspond to the underlined letters in Table \ref{tab:stats-rel}.}
\label{tab:perform-end-to-end}
\end{table}

\section{Computational Efficiency}

\begin{table*}[t!]
\centering
\begin{tabular}{|c|c|c|}
\hline
\multicolumn{1}{|l|}{\multirow{2}{*}{}} & \textbf{TB-Dense}   & \textbf{MATRES}     \\ \cline{2-3} 
Han et al. (2019)                       & -                   & 125.8 sent/s        \\
Han et al. (2020)                       & 325 sent/s          & -                   \\
\hline
\textbf{Graph-based Biaffine}           & \textbf{452 sent/s} & \textbf{386 sent/s} \\ \hline
\end{tabular}
\caption{Comparison of Relation Extraction Performance.}
\label{tab:performance}
\end{table*}

In Table \ref{tab:performance}, we report the sentence per second performance during inference stage of our model, and compare it to the baseline models. We re-implemented the system by Han et al. (2020) for benchmark. To establish a fair comparison, we removed both the score calculation and standard input/output portion during the evaluation process.
Every experiment is performed ten times on the TB-Dense dataset and taking the average.
We use a workstation equipped with AMD Ryzen 9 5900x and Nvidia GTX1060 6gb as our test bench.
For TB-Dense we trained it for 40 epochs, for MATRES we trained it for 19 epochs.

\section{Ablation Studies}

\begin{table*}[t!]
\centering
\begin{tabular}{|l|c|c|l|c|l|l|}
\hline
\multirow{3}{*}{}                  & \multicolumn{3}{c|}{\textbf{TB-Dense}}                        & \multicolumn{3}{c|}{\textbf{MATRES}}                                                            \\ \cline{2-7} 
                                   & \textbf{P}     & \textbf{R}    & \multicolumn{1}{c|}{\textbf{F1}}   & \textbf{P}            & \multicolumn{1}{c|}{\textbf{R}}    & \multicolumn{1}{c|}{\textbf{F1}}   \\
\hline
\multicolumn{1}{|c|}{- Biaffine Attention}  & 49.3 & 50.6 & 49.9 & \textbf{69.1} & 61.8 & 65.2 \\
\multicolumn{1}{|c|}{- ARC} & 49.5 & 50.5 & 50.0 & 68.1 & 63.8 & 65.9\\
\hline
\multicolumn{1}{|c|}{\textbf{All}} & \textbf{50.6} & \textbf{51.3} & \multicolumn{1}{c|}{\textbf{50.9}} & 65.8         & \multicolumn{1}{c|}{\textbf{66.8}} & \multicolumn{1}{c|}{\textbf{66.3}} \\ 
\hline
\end{tabular}
\caption{Performance of relation extraction: ablation studies on the major components of our model.}
\label{tab:ablation}
\end{table*}

In this section, we perform ablation tests to demonstrate the effect of each improvement.
Quantitative results can be found in Table \ref{tab:ablation}.

\subsection{Advantage of biaffine for label prediction}
We compared the label prediction result of our biaffine model to the modified model using two standard Linear layers. The result showed that the Biaffine Attention Mechanism provide slight improvement over Linear layers.
We also observe that our Biaffine Attention model runs faster on average.
\subsection{Advantage of two biaffine modules over one classification for NONE relation}
We use a special ARC module to predict the existence of an arc between two tokens.
In this section, we show how having a specialized ARC module in predicting NONE relations to be superior than taking the NONE relation as a label in the REL module.

In this experiment, we completely disable the ARC module and do not take an additional step to filter out NONE relations. 



\section{Analysis}
We analyze our experimental results from two aspects: The advantage of separate attention modules and the computational efficiency.

\subsection{Advantage of two biaffine attentions over one classification for NONE relation}

Most prior work uses NONE as a classification label in the temporal relation extraction process. In this work, however, the existence of an relation between two tokens is predicted directly by the ARC module. To show such improvement over traditional methods, we similarly added the NONE label to the REL module and completely disable the ARC module to compare the difference.

We found that without the ARC module, the overall F1 score of relation extraction decreases slightly, primary due to the degraded extraction performance on VAGUE labels. This proves that the detection of relation is better treated as a separated task, instead of being compared directly to other temporal relation labels.

\subsection{Computational Efficiency}

Benchmarks show that in a fixed amount of time, our model processes more sentences that previous studies, while providing comparable accuracy. This is primary due to our model design that removes the event extraction dependency for relation extraction. 
The prediction of temporal relation labels is performed in parallel with arc prediction, eliminating the need of the non-parallelizable, nested for-loop for extracting event pairs in the previous studies.

\section{Conclusions and Future Work}
In this paper, we propose a Graph-based Biaffine Attention Model for temporal relation extraction. We show that under end-to-end settings, our model predicts arcs and relation labels in the temporal graph directly, unlike standard two-step process that requires predictions of events before predicting relations. Our model is shown to provide better performance than traditional methods experimentally in F1-score and inference speed on benchmark datasets.

We plan to apply the proposed framework to other temporal reasoning tasks, such as extrating Time-Event and Time-Time relations.



\bibliography{anthology,emnlp2021}
\bibliographystyle{acl_natbib}

\appendix

\section{Hyperparameters}
\label{sec:appendix}

\begin{itemize}
    \item Sentence Encoder: 2-layer BiLSTM, hidden size=400 
    \item ARC module: We use two MLPs of same size, with output feature dimension=300
    \item REL module: We use two MLPs of same size, with output feature dimension=300
    \item Both BiLSTM and MLP dropout ratio=0.33
    \item Optimizer: lr=5e-5, mu=.9, nu=.9, epsilon=1e-12, clip=5.0, decay=.75
    \item All hyperparamters are tuned manually to optimize the development set in TB-Dense and MATRES respectively
\end{itemize}

\section{Dataset Summary}
\label{sec:appendix}
\begin{itemize}
    \item Please refer to Table \ref{tab:stats-rel} for relation label statistics and Table \ref{tab:splits} for document statistics
\end{itemize}

\section{Reproducibility List}
\label{sec:appendix}

\begin{itemize}
    \item All code and datasets can be found in the project code base.
    \item The pre-trained BERT-base-uncased model is provided by Huggingface transformers.
    \item Hyperparameters can be found in the Hyperparameters section.
\end{itemize}

\end{document}